\documentclass{article}

    \PassOptionsToPackage{numbers, compress}{natbib}
 \usepackage[preprint]{neurips_2025}

\usepackage[utf8]{inputenc} 
\usepackage[T1]{fontenc}    
\usepackage{hyperref}       
\usepackage{url}            
\usepackage{booktabs}       
\usepackage{amsfonts}       
\usepackage{nicefrac}       
\usepackage{microtype}      
\usepackage{xcolor}         

\usepackage{algpseudocode}
\usepackage{algorithm}
\usepackage{graphicx}
\usepackage{amsmath}
\usepackage{subcaption}

\title{Video CLIP Model for Multi-View Echocardiography Interpretation}

\author{%
    Ryo Takizawa$^{1, 2}$
    \quad
    Satoshi Kodera$^{1}$
    \quad
    Tempei Kabayama$^{1, 2}$
    \quad 
    Ryo Matsuoka$^{1}$ \\
    \textbf{Yuta Ando}$^{1}$
    \quad 
    \textbf{Yuto Nakamura}$^{1, 2}$
    \quad 
    \textbf{Haruki Settai}$^{1, 2}$
    \quad 
    \textbf{Norihiko Takeda}$^{1}$\\
    $^{1}$The University of Tokyo Hospital \quad $^{2}$The University of Tokyo \\
    \texttt{takizawa@isi.imi.i.u-tokyo.ac.jp}
}

\begin{document}
\maketitle

\begin{abstract}
Echocardiography records ultrasound videos of the heart, enabling clinicians to assess cardiac function. Recent advances in large-scale vision–language models (VLMs) have spurred interest in automating echocardiographic interpretation. However, most existing medical VLMs rely on single-frame (image) inputs, which can reduce diagnostic accuracy for conditions identifiable only through cardiac motion.
In addition, echocardiographic videos are captured from multiple views, each varying in suitability for detecting specific conditions. Leveraging multiple views may therefore improve diagnostic performance.
We developed a video–language model that processes full video sequences from five standard views, trained on 60,747 echocardiographic video–report pairs. We evaluated the gains in retrieval performance from video input and multi-view support, including the contributions of various pretrained models.
Code and model weights are available at \href{https://github.com/UTcardiology/video-echo-clip}{https://github.com/UTcardiology/video-echo-clip}.
\end{abstract}

\section{Introduction}
\label{sec:intro}

Echocardiography is a widely used, noninvasive method for diagnosing various cardiac conditions, including myocardial infarction, valvular diseases, and congenital heart defects. 
However, interpreting echocardiographic videos requires specialized expertise, which can be both time-consuming and costly, especially in emergency settings or areas lacking medical professionals. This has fueled growing interest in automated or AI-assisted diagnostic support.
Recent advances in VLMs have enabled the development of AI systems that interpret echocardiographic images at near-expert levels.
EchoCLIP \citep{Christensen2024} is a CLIP \citep{Radford2021} model trained on 1,032,975 echocardiographic images paired with clinical reports from 224,685 cases. 
By learning to align image embeddings with their corresponding report embeddings, EchoCLIP can assess disease presence and severity based on the inferred similarity between the images and reports. This CLIP-based approach provides a generalizable solution for interpreting diverse cardiac conditions. Furthermore, training vision encoders that effectively represent visual inputs is crucial for developing multimodal large language models (MLLMs) capable of generating clinical reports and comprehensive diagnoses.

Despite the progress made by EchoCLIP and other VLMs, two major challenges remain, given the unique nature of echocardiography: using videos instead of still images, and incorporating multiple views. 
Unlike static imaging methods such as chest X-rays, echocardiograms capture the heart’s rhythmic motion, an essential aspect for diagnosing certain conditions (e.g., valvular disease with abnormal blood flow).
Another key feature of echocardiography is its variety of views. Because the heart is a three-dimensional, anisotropic organ, positioning the ultrasound probe at different angles yields distinct cross-sections. While there are dozens of potential views, commonly used ones include the long-axis (LAX), short-axis (SAX), two-chamber (2CH), three-chamber (3CH), and four-chamber (4CH) views. Each view is especially useful for assessing specific aspects of cardiac function, indicating that further investigation is needed into performance improvements gained by integrating information from multiple views.

\begin{figure}[t!]\centering
\includegraphics[width=\linewidth]{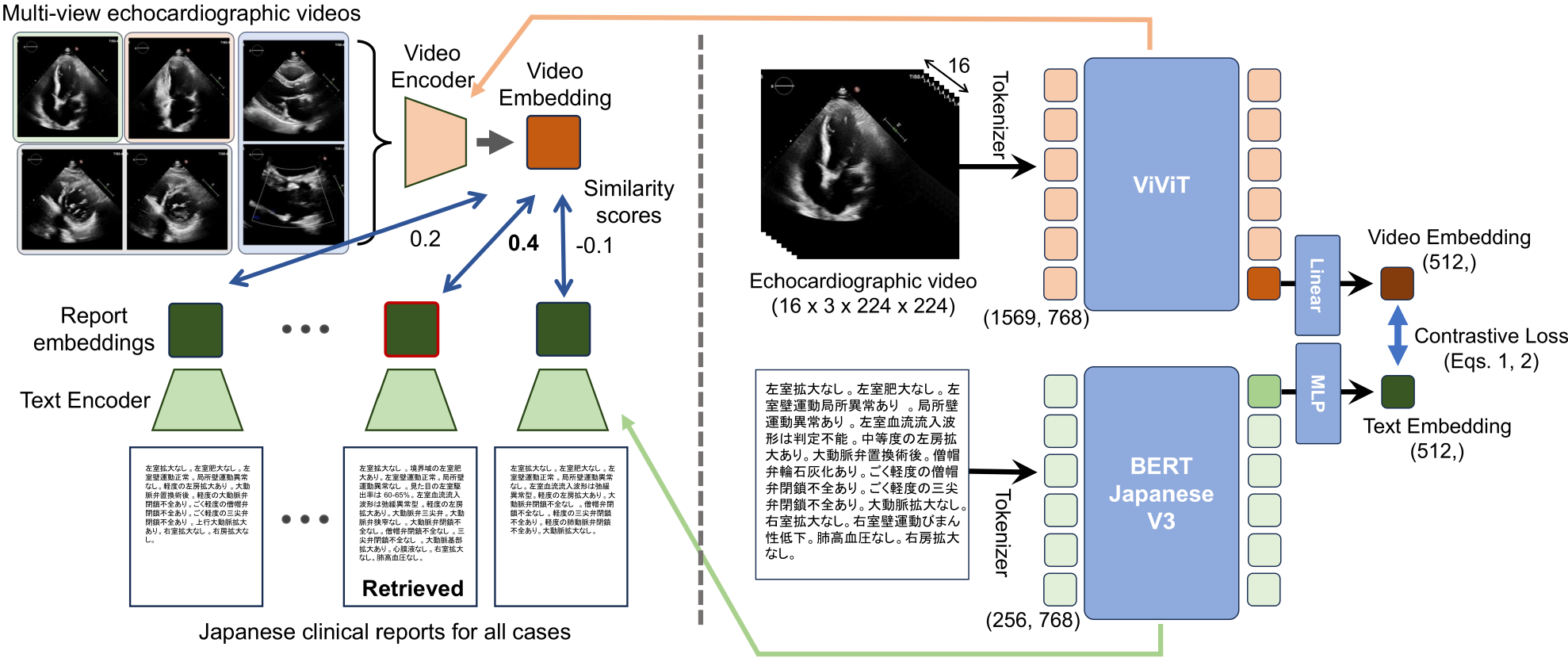}
\caption{Multi-view echocardiography interpretation (left) using video CLIP model (right). The most appropriate clinical report for the echocardiographic videos is retrieved by embedding similarity.}
\label{fig:overview}
\end{figure}

In this study, we aim to enhance the interpretation accuracy of a CLIP model by leveraging these two characteristics of echocardiography data (Fig.~\ref{fig:overview}). First, we replace the image encoder of a CLIP model with a video encoder~\citep{Arnab2021, Tong2022, Piergiovanni2022}, enabling the extraction of feature vectors that capture the temporal dynamics of echocardiogram videos. Second, we expand the dataset from the 4CH view to include five views---LAX, SAX, 2CH, 3CH, 4CH.
We train this model on a dataset containing 60,747 cases, comprising 747,900 pairs of multi-view echocardiogram videos and corresponding clinical reports from 29,886 patients. We then evaluate it by assessing its ability to retrieve the corresponding clinical reports from echocardiogram videos (video-to-text retrieval) and vice versa (text-to-video retrieval).

The recently proposed EchoPrime \citep{Vukadinovic2024} is a concurrent work that also extends a CLIP model to support multi-view and video inputs.
However, under controlled conditions, it does not examine the extent to which image-to-video or multi-view approaches actually contribute to interpretation performance.
In contrast, we train three ablation models to isolate the effects of video input and multi-view support during both training and inference. Moreover, we evaluate the impact of pretrained models for both video- and image-based settings.

\section{Method}
\label{sec:Method}

\subsection{Model Architecture}
In this study, following EchoCLIP, we perform contrastive learning on pairs of echocardiogram videos and their corresponding clinical reports, treating the correct (matching) video–report pairs as positive pairs and all others as negative pairs.
The overview of the model architecture is shown in Fig. \ref{fig:overview}. 
For the video encoder, we employ ViViT \citep{Tong2022}, which efficiently transforms a sequence of fixed-length frames (32 frames used) into a 512-dimensional embedding.
The text encoder used is BERT \citep{Devlin2019}.
Since the clinical reports are written in Japanese, we utilized BERTJapaneseV3 \citep{tohukuBERT}, which was pre-trained on a Japanese corpus.

Additionally, while CLIP models such as EchoCLIP typically use 77 tokens for the text encoder, clinical reports in echocardiography often describe each symptom and item in detail, necessitating longer text inputs. Therefore, we adopted 256 tokens for the text encoder.

\subsection{Multi-view Video Report Retrieval}
The video encoder and text encoder, trained through Contrastive Learning, are used to retrieve the most appropriate report from a set of candidate reports based on the similarity of embeddings during interpretation. 
For each symptom or item, its existence and severity (e.g., mild / moderate / severe) are associated with corresponding text, which is then converted into embeddings by the text encoder. The similarity between these text embeddings and the embedding of the target echocardiographic video is compared, and the text with the highest similarity is selected as the interpretation result.

However, in echocardiography, multiple echocardiographic videos from different views are taken for each case, and the physician creates a single report by comprehensively evaluating these videos. Similarly, in this study, all available echocardiographic videos for a given case are individually converted into embeddings, and their average is computed to obtain the overall video embedding. The similarity between this video embedding and the corresponding report embedding is then calculated and used to retrieve clinical reports. 
Reversely, it is also possible to retrieve the case with the most relevant echocardiographic videos for a given report.

\section{Experiments}
\label{sec:experiments}

\subsection{Dataset}
A total of 69,482 echocardiographic examination cases from 29,886 patients, collected between 2015 and 2023, were used to construct the dataset. 
These data were selected based on a separately trained view-classification CNN model, which assigned them to one of LAX, SAX, 2CH, 3CH, or 4CH views with a probability of at least 0.9. Any data classified into other views or assigned a lower probability were excluded beforehand.

\subsection{Ablation Models}
We evaluate the interpretative performance of the proposed multi-view video-input model (\textbf{MultiVideo}) by comparing it with two ablation models: a single-view video-input model (\textbf{SingleVideo}) and a single-view image-input model (\textbf{SingleImage}), the latter corresponding to EchoCLIP.

SingleVideo shares the same architecture as MultiVideo but is trained exclusively on 4CH-view videos. SingleImage replaces the video encoder with ConvNext-Base, an image encoder. The training dataset for SingleImage only includes the 4CH view, and a single frame randomly extracted from the video is used as input. To ensure a fair comparison, all models use the same text encoder and are trained from scratch.

For report retrieval, both SingleVideo and SingleImage use only 4CH-view videos as input. Unlike the video-based models, SingleImage computes the mean of all image embeddings extracted from every frame across the multiple videos, following the approach used in the EchoCLIP study.

\subsection{Model Comparison}

Table \ref{tab:tab2} shows the retrieval accuracy of the proposed model and the two ablation models. Accuracy is evaluated using mean cross-modal retrieval rank (MCMRR) and R@10. MCMRR represents the mean rank at which the correct report appears when all 5,515 reports are sorted by similarity, while R@$k$ indicates the percentage of cases where the correct report is ranked within the top $k$ positions.

\begin{table}[htbp]
    \centering
    \begin{tabular}{lcccc}
    \toprule
     & \multicolumn{2}{c}{MCMRR $\downarrow$} & \multicolumn{2}{c}{R@10 $\uparrow$} \\
    \cmidrule(lr){2-3} \cmidrule(lr){4-5}
    \textbf{Method} & V$\to$R & R$\to$V & V$\to$R & R$\to$V \\
    \midrule
    MultiVideo     & \textbf{595} & \textbf{584} & \textbf{10.9 \%} & \textbf{10.3 \%} \\
    MultiVideo-4CH & 705 & 695 & 8.4 \% & 8.0 \% \\
    SingleVideo    & 676 & 686 & 8.8 \% & 7.1 \% \\
    SingleImage    & 1222 & 1115 & 3.6 \% & 4.8 \% \\
    \bottomrule
    \end{tabular}
    \vspace{0.2cm}
    \caption{Retrieval scores for MultiVideo, SingleVideo and SingleImage (Video$\to$Report and Report$\to$Video).}
    \label{tab:tab2}
\end{table}

As shown in the table, the model with the highest readability performance was the multi-view video model (MultiVideo). The next highest was the 4CH-view-only video model (SingleVideo), followed by the 4CH-view-only image model (SingleImage). The most significant improvement in retrieval accuracy was observed when switching from image-based to video-based input, with both MCMRR and R@10 approximately doubling. Furthermore, incorporating multiple views led to an additional improvement of about 1.2 times.

To further evaluate the contribution of multi-view information, we also compared the performance of MultiVideo when restricted to the 4CH view at inference (MultiVideo-4CH) with that of SingleVideo. Their similar results suggest minimal knowledge transfer from multi-view training, indicating that the primary benefit of multi-view lies in providing diverse information during inference.

\subsection{Effect of Pretraining}
Beyond input modality and multi-view capability, the choice of pretrained models is a critical factor.
For models with image or video inputs, a key concern is that far stronger pretrained models are available for images.
To evaluate this effect, we trained SingleImage and MultiVideo-4CH models using different pretrained weights—ImageNet-21k, LAION-2B, Kinetics-400, and VideoMAE2—and compared retrieval performance. VideoMAE2 differs from ViViT in architecture and uses 16 frames per video.

\begin{figure}[htbp]
  \centering
  \begin{subfigure}[b]{0.45\textwidth}
    \centering
    \includegraphics[width=\textwidth]{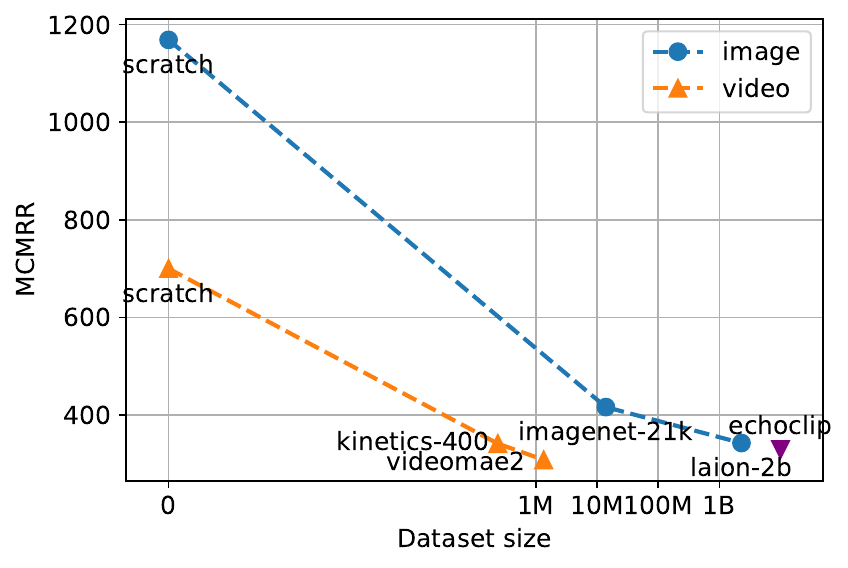}
  \end{subfigure}
  \hspace{0.7cm}
  \begin{subfigure}[b]{0.45\textwidth}
    \centering
    \includegraphics[width=\textwidth]{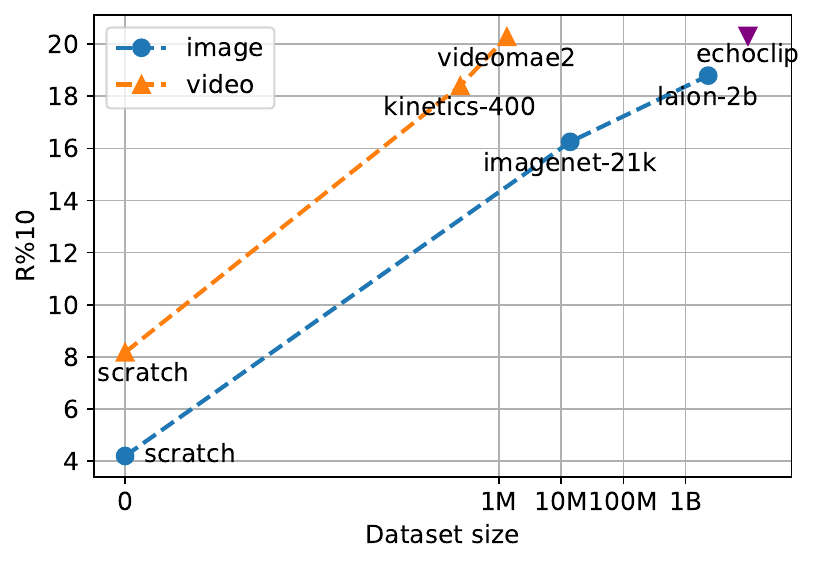}
  \end{subfigure}
\caption{Retrieval accuracy (log scale) of models trained with different pretrained initializations (left: MCMRR, right: R\%10).}
\label{fig:pretrain}
\end{figure}

As shown in Fig. \ref{fig:pretrain}, pretraining substantially improved performance for both image and video models.
Large-scale image pretraining greatly narrowed the performance gap between image and video models observed when both were trained from scratch.
However, the best performance was achieved by the VideoMAE2-based video model, surpassing even the EchoCLIP-pretrained image model.
Performance scaled almost linearly with video dataset size (log scale), suggesting that pretraining on 10M–100M-scale video datasets could yield further gains.
In contrast, image datasets showed gradual saturation, with diminishing returns expected even at multi-billion scale.

\section{Conclusion}
\label{sec:conclusion}
In this study, we focus on two key aspects of echocardiography: they ideally require video-based interpretation and they provide multiple views of the heart. 
Most VLM models applied to the medical domain so far have been single-image single-view approaches, so we extended these models to handle video inputs and multiple views for echocardiography. 
To assess the impact of these extensions, we compared the retrieval accuracy of the extended models with their unextended counterparts. 
The results show that, much like physicians, the CLIP model benefits from both video inputs and multi-view support. 

Because the healthcare field often restricts public data sharing, each organization’s accessible dataset tends to be limited. 
As a result, it becomes crucial to develop video-language models that maximize information extraction from available data. 
In the future, we plan to replace the vision encoder of existing MLLMs with the video encoder developed in this study and to build an echocardiography-specific MLLM capable of generating diagnostic reports.

\section*{Acknowledgement}
This work was supported by the Cross-ministerial Strategic Innovation Promotion Program (SIP) on “Integrated Health Care System” Grant Number JPJ012425.

{
    \small
    \bibliographystyle{plainnat}
    \bibliography{main}
}

\appendix

\section{Details on Report Retrieval}
The entire report retrieval process based on multi-view video interpretation is summarized in Algorithm~\ref{alg:alg1}.

\begin{algorithm}[htbp]
\caption{Report Retrieval from Multi-view Videos}
\label{alg:alg1}
\begin{algorithmic}[1]
    \Statex \textbf{Given:} $N$ reference videos from a single study $\{\phi_0, \phi_1, \ldots, \phi_N\}$, and reports from $M$ studies $\{\tau_0, \tau_1, \ldots, \tau_M\}$.
    \Statex \textbf{Notation:} Let $f(\cdot)$ and $g(\cdot)$ denote the trained video and text encoders, respectively.
    \vspace{0.1cm}
    \State $\{v_0, v_1, \ldots, v_N\} \leftarrow \{f(\phi_0), f(\phi_1), \ldots, f(\phi_N)\}$
    \State $v \leftarrow \text{mean}(v_0, v_1, \ldots, v_N)$ \Comment{average video embeddings}
    \State $s \leftarrow \emptyset$
    \For{$m = 1$ to $M$}
        \State $t_m \leftarrow g(\tau_m)$
        \State $s \leftarrow s \cup \frac{t_m^T v}{\|t_m\|\|v\|}$ \Comment{compute similarity}
    \EndFor
    \Statex \textbf{Return:} $\tau_{\text{argmax}(s)}$ \Comment{retrieved report}
\end{algorithmic}
\end{algorithm}

\section{Training Details}
\subsection{Contrastive Learning Loss}
For a batch of size \( B \) containing pairs of echocardiogram videos and clinical reports, we obtain embeddings \( \{(v_i, t_i)\}_{i=0, \ldots, B} \) using the video encoder and text encoder, respectively. The contrastive loss can then be expressed as follows:
\begin{align}
    &\mathcal{L}_{\text{video-to-report}} = \frac{1}{B}\sum_{i=0}^B-\log\frac{\exp\left(\frac{1}{\tau}\frac{t_i^Tv_i}{\|t_i\|\|v_i\|}\right)}{\sum_{i}\exp\left(\frac{1}{\tau}\frac{t_i^Tv_i}{\|t_i\|\|v_i\|}\right)}, \label{eq:loss-video-to-text}\\
    &\mathcal{L}_{\text{report-to-video}} = \frac{1}{B}\sum_{i=0}^B-\log\frac{\exp\left(\frac{1}{\tau}\frac{v_i^Tt_i}{\|v_i\|\|t_i\|}\right)}{\sum_{i}\exp\left(\frac{1}{\tau}\frac{v_i^Tt_i}{\|v_i\|\|t_i\|}\right)},   \label{eq:loss-text-to-video}
\end{align}
where $\tau$ denotes temperature.
Eq.~\eqref{eq:loss-video-to-text} represents the contrastive loss for video-to-report, while Eq.~\eqref{eq:loss-text-to-video} represents the contrastive loss for report-to-video. The training loss is the average of both.

\subsection{Dataset}
Table \ref{tab:tab1} summarizes the dataset. 
The patients were split into training, validation, and test sets in a ratio of 0.875:0.025:0.1. 
MultiVideo was trained on 747,900 multi-view videos, whereas SingleVideo and SingleImage were trained on 184,444 4CH-view videos. 
For the test set, in order to compare 4CH-view and multi-view approaches, 5,515 of the 7,050 cases that contained a 4CH-view video were used.

\begin{table}[htbp]
    \centering
    \begin{tabular}{lccc}
    \toprule
                    & Train   & Valid   & Test \\
    \midrule
    Case            & 60,747  & 1,685   & 7,050 (5,515) \\
    Patient         & 29,886  & 853     & 3,416 (2,917) \\
    LAX-view Video  & 201,253 & 5,758   & 23,358 \\
    SAX-view Video  & 191,577 & 5,477   & 22,068 \\
    2CH-view Video  & 65,630  & 1,777   & 7,405 \\
    3CH-view Video  & 104,996 & 2,915   & 12,062 \\
    4CH-view Video  & 184,444 & 5,113   & 21,345 \\
    Total Video     & 747,900 & 21,040  & 86,238 (78,276) \\
    \bottomrule
    \end{tabular}
    \vspace{0.5cm}
    \caption{Summary of the dataset. The values in parentheses indicate cases that include 4CH-view videos.}
    \label{tab:tab1}
\end{table}

\subsection{Training}
All models were trained on four NVIDIA H100 GPUs. A batch size of 64 was used for the video-based models and 2,304 for the image-based model. The learning rate was set to 1e-5, with a linear warm-up during the first 2,000 steps followed by a cosine-annealing schedule. Training each model required one to two days.

\section{Example of Data and Retrieval Results}
\begin{figure}[htbp]\centering
\includegraphics[width=\textwidth]{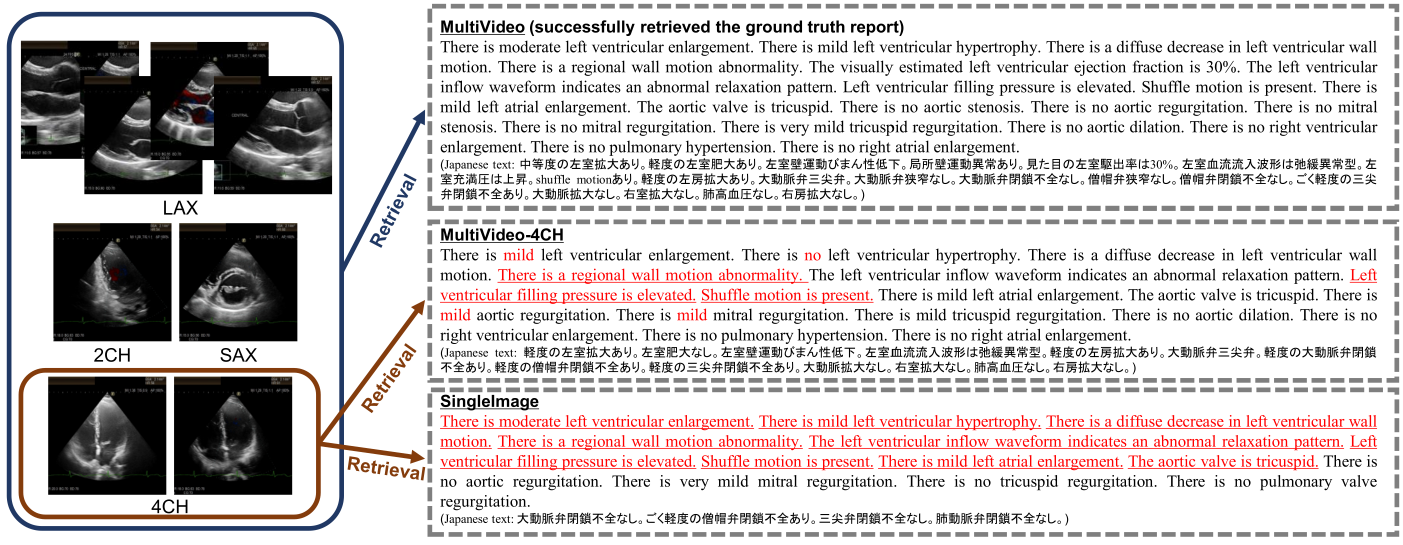}
\caption{An example of the most similar clinical reports retrieved from 5,515 candidates for a specific echocardiogram case by SingleImage, MultiVideo-4CH, and MultiVideo. Text in red denotes discrepancies from the ground truth, while underlined text indicates missing content.}
\label{fig:retrieved-report-results}
\end{figure}

In Figure \ref{fig:retrieved-report-results}, one can see an example of the echocardiographic videos and corresponding reports (the ground truth and retrieved) used in the experiments. 
Each study contains videos from various views, and the number of views and videos per view differs across cases. 
Clinical reports describe whether symptoms are present and to what degree. Any text exceeding 256 tokens was truncated.

These three clinical reports are considered most similar out of 5,515 possible reports by MultiVideo, MultiVideo-4CH, and SingleImage for the given echocardiogram video / image. 
In this case, the discrepancy (in red text in the figure) between the retrieved clinical reports and the ground truth report decreases in the order of SingleImage, MultiVideo-4CH, and then MultiVideo.
The decline in left ventricular systolic function, difficult to assess from still images, was not detected by SingleImage, yet it was correctly interpreted by the video-based models, MultiVideo-4CH and MultiVideo. 
Furthermore, for conditions such as left ventricular enlargement and hypertrophy, which are difficult to identify using only the 4CH view, MultiVideo was more accurate than SingleImage or MultiVideo-4CH.

\end{document}